\documentclass{article}

\usepackage[numbers]{natbib} 

\usepackage[preprint]{neurips_2023}

\usepackage[utf8]{inputenc} 
\usepackage[T1]{fontenc}    
\usepackage{hyperref}       
\usepackage{url}            
\usepackage{booktabs}       
\usepackage{amsfonts}       
\usepackage{nicefrac}       
\usepackage{microtype}      
\usepackage{xcolor}         
\usepackage{graphicx}
\usepackage{multirow}
\usepackage{float}

\title{Natively Trainable Sparse Attention \\for Hierarchical Point Cloud Datasets}

\author{%
  Nicolas Lapautre\textsuperscript{*} \\
  University of Groningen\\
  \texttt{n.a.lapautre@student.rug.nl} \\
  \And
  Maria Marchenko\textsuperscript{*} \\
  University of Amsterdam\\
  \texttt{maria.marchenko@student.uva.nl} \\
  \And
  Carlos Miguel Patiño\textsuperscript{*} \\
  University of Amsterdam\\
  \texttt{carlos.patino.paz@student.uva.nl} \\
  \And
  Xin Zhou\textsuperscript{*} \\
  University of Amsterdam\\
  \texttt{xin.zhou@student.uva.nl} \\
}

\begin{document}

\maketitle

\begin{abstract}
Unlocking the potential of transformers on datasets of large physical systems depends on overcoming the quadratic scaling of the attention mechanism. This work explores combining the Erwin architecture with the Native Sparse Attention (NSA) mechanism to improve the efficiency and receptive field of transformer models for large-scale physical systems, addressing the challenge of quadratic attention complexity. We adapt the NSA mechanism for non-sequential data, implement the Erwin NSA model, and evaluate it on three datasets from the physical sciences---cosmology simulations, molecular dynamics, and air pressure modeling---achieving performance that matches or exceeds that of the original Erwin model. Additionally, we reproduce the experimental results from the Erwin paper to validate their implementation.
\end{abstract}

\section{Introduction}

Self-attention has emerged as a fundamental mechanism in deep learning, revolutionizing how models capture complex relationships within data in areas like natural language processing (NLP), computer vision (CV), and 3D point clouds. The most significant benefit of self-attention is that it enables capturing both local and long-range dependencies by allowing information flow between all tokens or nodes. However, using self-attention in scenarios that require modeling interactions between a large number of tokens or nodes scales poorly because of the quadratic complexity of the mechanism.

Our work addresses the quadratic scaling in self-attention through model architecture and hardware-aware optimizations for tasks related to point clouds in the physical sciences. We build upon the hierarchical transformer proposed in Erwin \cite{zhdanov2025erwin} and enhance its efficiency through the Native Sparse Attention (NSA) mechanism \cite{yuan2025native}. Erwin is a hierarchical transformer that achieves sub-quadratic scaling while maintaining state-of-the-art (SOTA) performance in point cloud datasets that require modeling short- and long-scale interactions. However, Erwin relies on a U-Net architecture to capture long-range interactions, so the strength of the receptive field is bottlenecked by the pooling operations in the U-Net layers.

NSA integrates algorithmic innovations and hardware-aligned optimizations by introducing three attention mechanisms---compressed, selection, and sliding attentions---that achieve accelerated training and inference while maintaining the performance of a complete attention mechanism. The NSA was initially proposed to capture long- and short-term relationships for sequential tasks, but we extend their idea to work on 3D point clouds. By combining Erwin's architecture with Native Sparse Attention, an approach we refer to as Erwin NSA, we leverage the speedup observed from NSA without compromising model performance in sequential tasks. We evaluate our approach using three datasets used by \citet{zhdanov2025erwin} to determine how Erwin NSA compares to Erwin in model performance, runtime, and memory consumption.

\section{Related work}

\subsection{Sub-quadratic scaling for attention}

Several works have addressed attention's quadratic scaling by proposing methods such as locality-sensitivity hashing \cite{Kitaev2020ReformerTEA}, sparse attention \cite{Zaheer2020BigBTA}, and low-rank approximation \cite{Wang2020LinformerSWA} in the context of NLP, as well as non-overlapping patches \cite{liu2021swin} in the context of CV. These methods leverage the assumption that data is regularly distributed across a grid---i.e., sequences of tokens in NLP and grids of pixels in CV. This regular distribution on a grid is not proper for applications like 3D point clouds and non-uniform meshes, so other works have explored approaches like converting point clouds into sequences \cite{wu2024point}, hierarchical attention \cite{zhu2021h,kang2023fast,wang2023octformer}, and cluster attention \cite{janny2023eagle,alkin2024universal}.

Beyond model innovations, work like FlashAttention \cite{shah2024flashattention3fastaccurateattention} speeds up model training and inference without a compromise in model performance by accounting for how GPUs handle memory operations during the attention operation. This highlights the importance of not only focusing on efficient model architectures but also identifying opportunities to incorporate implementations of existing architectures that optimize runtime and memory usage by leveraging hardware capabilities effectively.

\subsection{Erwin architecture}
The Erwin architecture is a hierarchical transformer designed for large-scale physical systems. It leverages ball tree partitioning to organize computation on a point cloud, enabling linear-time attention by processing nodes in parallel within local balls. Erwin captures both fine-grained local details and global context through a U-net architecture which progressively coarsens and refines the ball tree structure. Additionally, cross-ball interaction is further captured  by rotating the ball tree. 
\subsection{NSA mechanism}

 Native Sparse Attention (NSA) is a novel attention mechanism \cite{yuan2025native} designed for long-context modeling in language models which integrates hierarchical token modeling with sparse attention. It reduces per-query computation by organizing keys and values into temporal blocks. These blocks are processed through three attention paths—compressed coarse-grained tokens, fine-grained token blocks, and sliding window attention—which are then concatenated to form the final output. The selection of fine-grained blocks is guided by importance scores derived from the compressed token attention. The final output is a learned combination of these three attention branches. NSA is shown to maintain the accuracy of Full Attention while achieving a speedup of 2x to 9x, depending on the context length, with greater speedups for longer context lengths.
 
\section{Reproducibility}
To validate the Erwin Transformer's capability to capture long-range interactions in point clouds, accelerate simulations, and enhance expressivity, we replicate three experiments from the original paper: cosmological simulations, molecular dynamics, and airflow pressure modeling.
\subsection{Experiments}
\textbf{Cosmological simulations}: The cosmology dataset \cite{balla2024cosmic} contains large-scale point clouds representing galaxy distributions. The input is a point cloud $X\in \mathbb{R}^{5000\times 3}$ , where each row represents a galaxy with $x, y, z$ coordinates. The model predicts the velocity of each galaxy $Y\in \mathbb{R}^{5000\times 3}$. The evaluation metric is the mean squared error (MSE) between predicted and ground truth velocities.

\textbf{Molecular dynamics (MD)}: The MD dataset \cite{webb2020targeted, fu2022simulate} consists of single-chain polymers. The model input is a polymer chain of $N$ beads, each with a specific weight and associated velocities from the previous 16 timesteps. The model predicts the mean and variance of acceleration for each bead. The results are evaluated using negative log-likelihood (NLL).

\textbf{Airflow pressure modeling}: The ShapeNet-Car dataset \cite{umetani2018learning, alkin2024universal} consists of 889 car models, each represented by 3586 3D surface points $X\in \mathbb{R}^{3586\times 3}$. The model predicts the pressure $P\in \mathbb{R}^{3586\times 1}$ at each point, evaluated using mean squared error (MSE) between predicted and ground truth pressures.

The three experiments are replicated with the same model configurations and hyperparameters as in the original paper, which can also be found in Appendix \ref{sec:appendix reproduce}, and compared to its recorded results.

\subsection{Results and discussion}
As shown in Table \ref{table:replicate}, the replicated results closely match the original ones. The original ShapeNet result is from 1000 epochs, not after convergence, and the difference may be due to early training fluctuations. After contacting the author for the convergence result and replicating the training, both achieved an MSE ($\times 10^{-2}$) of 0.16.

\begin{table}[H] 
\centering
\begin{tabular}{@{}llcc@{}}
\toprule
Dataset & Metric & Original Results & Replicated Results\\
\midrule
Cosmology & Test MSE, $\times10^{-2}$ & 0.60 & 0.60 \\
\midrule
MD & Test NLL & 0.71 & 0.70 \\
\midrule
ShapeNet & Test MSE, $\times10^{-2}$ & 1.42 & 1.30\\
\bottomrule
\end{tabular}
    \caption{Summary of replication results compared to the original. The results align closely for Cosmology and MD datasets, while the difference in ShapeNet may be due to early training fluctuations, as the results were recorded before convergence.}
    \label{table:replicate}
\end{table}

\section{Extension}
Our extension involves combining the Erwin architecture with the NSA mechanism. NSA's $O(n \sqrt n)$ complexity enables global attention even in large point clouds. We hypothesize that NSA expands the nodes’ receptive field, improving performance while maintaining Erwin’s efficiency.

\subsection{Methodology}

\begin{figure}[H]
    \centering
    \begin{minipage}{0.37\textwidth}
        \centering
        \includegraphics[width=\linewidth]{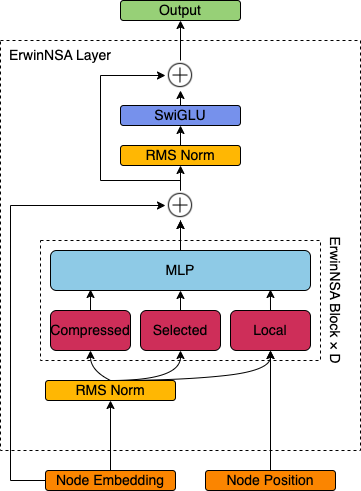}
        \label{fig:architecture1}
    \end{minipage}%
    \hspace{0.5cm}  
    \begin{minipage}{0.45\textwidth}
        \centering
        \includegraphics[width=\linewidth]{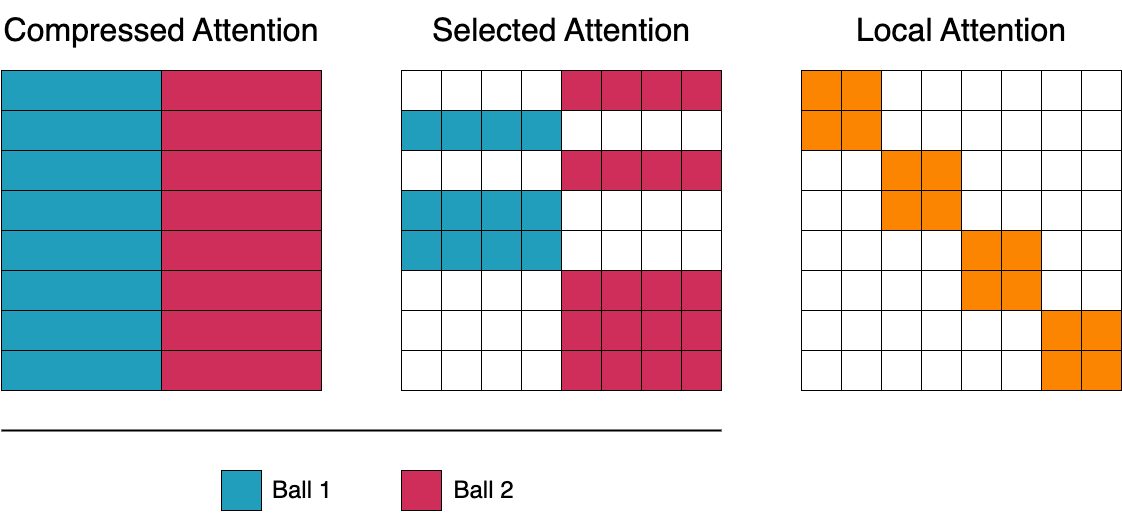}
        \label{fig:architecture2}
    \end{minipage}
    \caption{Erwin NSA architecture (left) and details of the three attention mechanisms we use for our NSA implementation (right). The compressed and selected attention mechanisms are the same as the original NSA implementation. We replace the sliding window attention from NSA with a local attention mechanism to learn local patterns in the context of a hierarchical ball structure.}
    \label{fig:architecture}
\end{figure}

Since NSA is global by nature, we remove Erwin's U-net coarsening and refinement architecture and utilize Erwin's tree partitioning to organize global cross-ball interaction. We adapt the NSA method for non-sequential data by redefining its attention components to operate over a hierarchical ball-tree structure. At each tree level, we compress the representations within each ball. Then, for every point in a ball, we compute attention scores over all compressed balls, select the top-k based on these scores, and apply full attention between the point and all leaf nodes in the selected balls (see right part of Figure \ref{fig:architecture} for visualization). These steps correspond to the token compression and selection mechanisms in NSA.

We argue that sliding window attention is not applicable in our case, as our data is not sequential. Instead, we propose using local ball attention, motivated as follows: in sequential data, a sliding window captures the nearest causal tokens. In the context of a ball-tree layer, the spatial neighborhood of a point is approximated by the ball that contains it. Therefore, attention within this ball serves a similar role to sliding window attention, capturing local patterns. For this local attention, we use Erwin's relative positional embedding and attention bias.

As in Erwin, we use an MPNN to obtain initial node embeddings and reuse rotations defined in Erwin. Instead of the U-net encoder-decoder architecture, we fix the compressed and local ball sizes for each of the NSA blocks, which are composed sequentially. Our NSA layer includes two residual connections. For details, refer to Figure \ref{fig:architecture}.

\subsection{Implementation}

We began by combining lucidrains' NSA implementation with Erwin's version. However, when testing on ShapeNet-Car, this initial implementation consumed 60GB of VRAM to train a network of the same size as Erwin's, which only required 2GB. To diagnose this issue, we used the PyTorch profiler utility. After identifying the memory bottlenecks, we optimized our code, including reimplementing local ball attention using PyTorch operations and adapting a selected Triton kernel for attention from \cite{fla_native_sparse_attention_2025}. Our implementation is available on GitHub\footnote{https://github.com/VZcosmos/ErwinNSA}. 

\begin{figure}[H]
    \centering
    \includegraphics[width=0.8\linewidth]{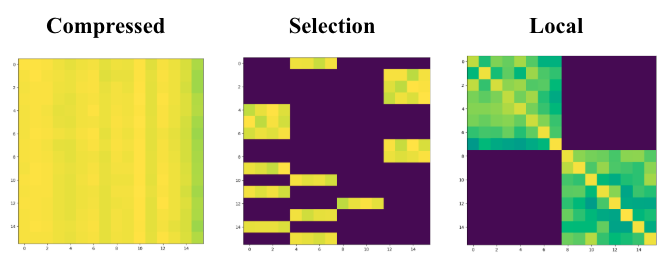}
    \caption{Visualization of the key-query matrices for each of the branches of NSA for a toy example of a sequence of 16 nodes. The plot in the left shows how all keys can attend to the query through the compressed mechanism; the plot in the middle how the patterns looks if we choose a block size of four and select the top block; the plot on the right how the nodes can attend to other nodes in a ball with size 8.}
    \label{fig:influence}
\end{figure}

\subsection{Experiments}
We compare the computational benchmark (memory usage, steps per second), node interactions (gradients), and task-specific performance (mean-squared error) across three datasets used in Erwin's paper: cosmology, molecular dynamics, and Shape-Net Car. All experiments are run on a batch size of one, due to limitations of our implementation. For each dataset, we chose individual hyperparameters, which we specify in Table \ref{tab:hyperparameters_comparison}.

\subsection{Results and Discussion}

We implemented and tested the Erwin NSA architecture. In our experiments, we were able to match---and in two datasets, outperform---Erwin by using a shallower Erwin NSA transformer with a higher number of steps per second. While Erwin and Erwin NSA influence the same number of nodes, the influence values in Erwin NSA are higher (see Figure \ref{fig:influence}), enabling smoother information flow and making Erwin NSA more expressive.

\begin{figure}[H]
    \centering
    \includegraphics[width=0.8\linewidth]{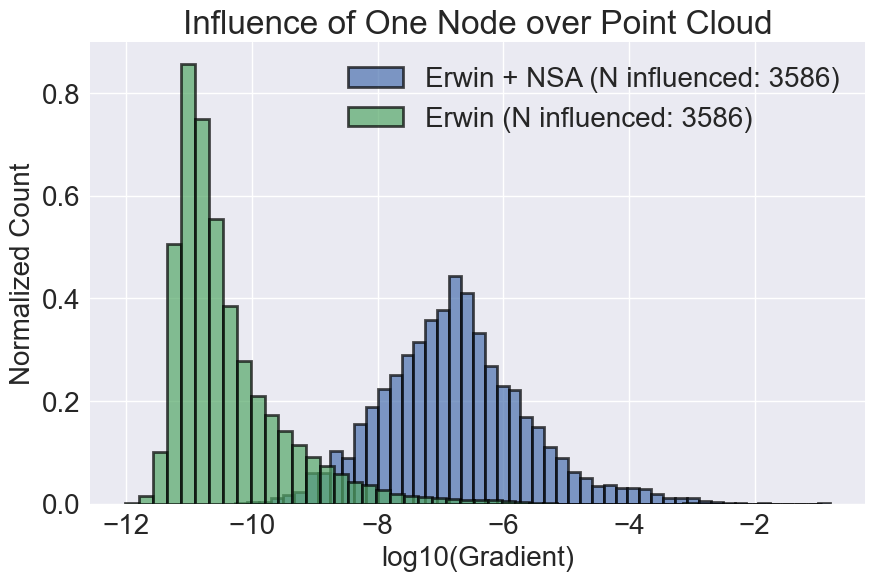}
    \caption{Comparison of the gradients of the output node with respect to an input node, where a higher value means the input node has a higher influence over the output node. As shown by the two distributions, both architectures have the same number of influenced points; however, the Erwin architecture with NSA allows a higher influence of an input node over the model's output on a specific node.}
    \label{fig:influence}
\end{figure}

Another advantage of our implementation is that it avoids the complexity of Erwin’s configuration by not relying on a U-Net architecture, which involves more hyperparameters and architectural decisions. Additionally, our approach addresses a key limitation of U-Net: its rigid, fixed-scale hierarchy, which restricts the ability to capture adaptive multi-scale interactions.

\begin{table}[H]
    \centering
    \begin{tabular}{@{}llcc@{}}
\toprule
Dataset       & Metric            & Erwin            & Erwin NSA Ours           \\ \midrule
\multirow{3}{*}{Cosmology} 
              & Test MSE          & \textbf{0.62}    & 0.63             \\
              & Steps per Second  & 3.47    & \textbf{3.79}             \\
              & Peak GPU Memory   & \textbf{1.01}    & 2.63             \\ \midrule
\multirow{3}{*}{MD} 
              & Test MSE          & 0.34    & \textbf{0.32}             \\
              & Steps per Second  & 14.64   & \textbf{36.58}            \\
              & Peak GPU Memory   & \textbf{0.35}    & 0.36             \\
              \midrule
\multirow{3}{*}{ShapeNet} 
              & Test MSE          & 74.40            & \textbf{20.26}   \\
              & Steps per Second  & 8.06             & \textbf{9.34}    \\
              & Peak GPU Memory   & \textbf{1.31}    & \textbf{1.31}    \\ 
              \bottomrule
\end{tabular}
    \caption{Summary of all the evaluations comparing the original Erwin implementation with our version, including NSA. Receptive field size is the same for Erwin and Erwin NSA, but the gradient distribution is different (See Figure \ref{fig:influence}).}
    \label{tab:all_evaluations}
\end{table}

We can identify a few future directions. A detailed ablation study can guide further optimizations. For example, we can try to remove compressed attention from the weighted values sum, and only use its similarity matrix for the selection attention. Furthermore, we can optimize this selection by compressing queries as well, allowing this part of NSA to be linear or even sublinear. We can explore other clustering algorithms to obtain the balls, as we do not need a tree but only a partitioning for compressed and local attentions. This way we can include priors about the data into the model (e.g. use graph-based clustering for molecular dynamics dataset).

\section{Conclusion}
We successfully reproduced the benchmark results of the three datasets in Erwin's paper. Following this, we modified NSA's implementation for sequential data to effectively integrate in ball tree partitioning over point clouds, while also optimizing it's memory usage and runtime.

Our combined Erwin NSA model outperforms Erwin on the ShapeNet and Molecular Dynamics datasets, and achieves comparable results on the Cosmology dataset.

\newpage

\small

\bibliography{references}

\begin{thebibliography}{18}
\providecommand{\natexlab}[1]{#1}
\providecommand{\url}[1]{\texttt{#1}}
\expandafter\ifx\csname urlstyle\endcsname\relax
  \providecommand{\doi}[1]{doi: #1}\else
  \providecommand{\doi}{doi: \begingroup \urlstyle{rm}\Url}\fi

\bibitem[Zhdanov et~al.(2025)Zhdanov, Welling, and van~de Meent]{zhdanov2025erwin}
Maksim Zhdanov, Max Welling, and Jan-Willem van~de Meent.
\newblock Erwin: A tree-based hierarchical transformer for large-scale physical systems.
\newblock \emph{arXiv preprint arXiv:2502.17019}, 2025.

\bibitem[Yuan et~al.(2025)Yuan, Gao, Dai, Luo, Zhao, Zhang, Xie, Wei, Wang, Xiao, et~al.]{yuan2025native}
Jingyang Yuan, Huazuo Gao, Damai Dai, Junyu Luo, Liang Zhao, Zhengyan Zhang, Zhenda Xie, YX~Wei, Lean Wang, Zhiping Xiao, et~al.
\newblock Native sparse attention: Hardware-aligned and natively trainable sparse attention.
\newblock \emph{arXiv preprint arXiv:2502.11089}, 2025.

\bibitem[Kitaev et~al.(2020)Kitaev, Kaiser, and Levskaya]{Kitaev2020ReformerTEA}
Nikita Kitaev, Lukasz Kaiser, and Anselm Levskaya.
\newblock Reformer: The efficient transformer.
\newblock \emph{ArXiv}, abs/2001.04451, 2020.
\newblock URL \url{https://arxiv.org/pdf/2001.04451.pdf}.

\bibitem[Zaheer et~al.(2020)Zaheer, Guruganesh, Dubey, Ainslie, Alberti, Onta{\~n}{\'o}n, Pham, Ravula, Wang, Yang, and Ahmed]{Zaheer2020BigBTA}
M.~Zaheer, Guru Guruganesh, Kumar~Avinava Dubey, J.~Ainslie, Chris Alberti, Santiago Onta{\~n}{\'o}n, Philip Pham, Anirudh Ravula, Qifan Wang, Li~Yang, and Amr Ahmed.
\newblock Big bird: Transformers for longer sequences.
\newblock \emph{ArXiv}, abs/2007.14062, 2020.
\newblock URL \url{https://arxiv.org/pdf/2007.14062.pdf}.

\bibitem[Wang et~al.(2020)Wang, Li, Khabsa, Fang, and Ma]{Wang2020LinformerSWA}
Sinong Wang, Belinda~Z. Li, Madian Khabsa, Han Fang, and Hao Ma.
\newblock Linformer: Self-attention with linear complexity.
\newblock \emph{ArXiv}, abs/2006.04768, 2020.
\newblock URL \url{https://arxiv.org/pdf/2006.04768.pdf}.

\bibitem[Liu et~al.(2021)Liu, Lin, Cao, Hu, Wei, Zhang, Lin, and Guo]{liu2021swin}
Ze~Liu, Yutong Lin, Yue Cao, Han Hu, Yixuan Wei, Zheng Zhang, Stephen Lin, and Baining Guo.
\newblock Swin transformer: Hierarchical vision transformer using shifted windows.
\newblock In \emph{Proceedings of the IEEE/CVF international conference on computer vision}, pages 10012--10022, 2021.

\bibitem[Wu et~al.(2024)Wu, Jiang, Wang, Liu, Liu, Qiao, Ouyang, He, and Zhao]{wu2024point}
Xiaoyang Wu, Li~Jiang, Peng-Shuai Wang, Zhijian Liu, Xihui Liu, Yu~Qiao, Wanli Ouyang, Tong He, and Hengshuang Zhao.
\newblock Point transformer v3: Simpler faster stronger.
\newblock In \emph{Proceedings of the IEEE/CVF Conference on Computer Vision and Pattern Recognition}, pages 4840--4851, 2024.

\bibitem[Zhu and Soricut(2021)]{zhu2021h}
Zhenhai Zhu and Radu Soricut.
\newblock H-transformer-1d: Fast one-dimensional hierarchical attention for sequences.
\newblock \emph{arXiv preprint arXiv:2107.11906}, 2021.

\bibitem[Kang et~al.(2023)Kang, Tran, and De~Sterck]{kang2023fast}
Yanming Kang, Giang Tran, and Hans De~Sterck.
\newblock Fast multipole attention: A divide-and-conquer attention mechanism for long sequences.
\newblock \emph{arXiv preprint arXiv:2310.11960}, 2023.

\bibitem[Wang(2023)]{wang2023octformer}
Peng-Shuai Wang.
\newblock Octformer: Octree-based transformers for 3d point clouds.
\newblock \emph{ACM Transactions on Graphics (TOG)}, 42\penalty0 (4):\penalty0 1--11, 2023.

\bibitem[Janny et~al.(2023)Janny, Beneteau, Nadri, Digne, Thome, and Wolf]{janny2023eagle}
Steeven Janny, Aur{\'e}lien Beneteau, Madiha Nadri, Julie Digne, Nicolas Thome, and Christian Wolf.
\newblock Eagle: Large-scale learning of turbulent fluid dynamics with mesh transformers.
\newblock \emph{arXiv preprint arXiv:2302.10803}, 2023.

\bibitem[Alkin et~al.(2024)Alkin, F{\"u}rst, Schmid, Gruber, Holzleitner, and Brandstetter]{alkin2024universal}
Benedikt Alkin, Andreas F{\"u}rst, Simon Schmid, Lukas Gruber, Markus Holzleitner, and Johannes Brandstetter.
\newblock Universal physics transformers: A framework for efficiently scaling neural operators.
\newblock \emph{Advances in Neural Information Processing Systems}, 37:\penalty0 25152--25194, 2024.

\bibitem[Shah et~al.(2024)Shah, Bikshandi, Zhang, Thakkar, Ramani, and Dao]{shah2024flashattention3fastaccurateattention}
Jay Shah, Ganesh Bikshandi, Ying Zhang, Vijay Thakkar, Pradeep Ramani, and Tri Dao.
\newblock Flashattention-3: Fast and accurate attention with asynchrony and low-precision, 2024.
\newblock URL \url{https://arxiv.org/abs/2407.08608}.

\bibitem[Balla et~al.(2024)Balla, Mishra-Sharma, Cuesta-Lazaro, Jaakkola, and Smidt]{balla2024cosmic}
Julia Balla, Siddharth Mishra-Sharma, Carolina Cuesta-Lazaro, Tommi Jaakkola, and Tess Smidt.
\newblock A cosmic-scale benchmark for symmetry-preserving data processing.
\newblock \emph{arXiv preprint arXiv:2410.20516}, 2024.

\bibitem[Webb et~al.(2020)Webb, Jackson, Gil, and de~Pablo]{webb2020targeted}
Michael~A Webb, Nicholas~E Jackson, Phwey~S Gil, and Juan~J de~Pablo.
\newblock Targeted sequence design within the coarse-grained polymer genome.
\newblock \emph{Science advances}, 6\penalty0 (43):\penalty0 eabc6216, 2020.

\bibitem[Fu et~al.(2022)Fu, Xie, Rebello, Olsen, and Jaakkola]{fu2022simulate}
Xiang Fu, Tian Xie, Nathan~J Rebello, Bradley~D Olsen, and Tommi Jaakkola.
\newblock Simulate time-integrated coarse-grained molecular dynamics with multi-scale graph networks.
\newblock \emph{arXiv preprint arXiv:2204.10348}, 2022.

\bibitem[Umetani and Bickel(2018)]{umetani2018learning}
Nobuyuki Umetani and Bernd Bickel.
\newblock Learning three-dimensional flow for interactive aerodynamic design.
\newblock \emph{ACM Transactions on Graphics (TOG)}, 37\penalty0 (4):\penalty0 1--10, 2018.

\bibitem[{FLA Organization}(2025)]{fla_native_sparse_attention_2025}
{FLA Organization}.
\newblock Native sparse attention.
\newblock \url{https://github.com/fla-org/native-sparse-attention}, 2025.
\newblock Git commit bd67af5, accessed 2025-05-31.

\end{thebibliography}
\bibliographystyle{unsrtnat}

\appendix
\section*{Appendix}

\section{Hyperparameters for comparison of Erwin and Erwin NSA}

\begin{table}[H]
    \centering
    \begin{tabular}{@{}llc@{}}
\toprule
Dataset       & Hyperparameter            & Value            \\ \midrule
\multirow{6}{*}{Cosmology} 
              & Local Ball Size           & 128     \\
              & Compressed Ball Size      & 32              \\
              & Number of Selected Balls  & 16     \\
              & Number of Epochs          & 10000                \\
              & Depth                     & 4                \\
              & \texttt{c\_hidden}        & 64                \\ \midrule
\multirow{6}{*}{MD} 
              & Local Ball Size           & 32              \\
              & Compressed Ball Size      & 32             \\
              & Number of Selected Balls  & 16     \\
              & Number of Epochs          & 90000                \\
              & Depth                     & 2                \\
              & \texttt{c\_hidden}        & 128              \\ \midrule
\multirow{6}{*}{ShapeNet} 
              & Local Ball Size           & 128             \\
              & Compressed Ball Size      & 32              \\
              & Number of Selected Balls  & 16     \\
              & Number of Epochs          & 80000                \\
              & Depth                     & 6                \\
              & \texttt{c\_hidden}        & 64                \\
              \bottomrule
\end{tabular}
    \caption{Summary of all hyperparameters used for comparison of Erwin NSA and Erwin. For Erwin, we used the default "small" configuration, with batch size 1 and the same number of epochs as Erwin NSA. Hyperparameters that are absent from the table are set to the same values as in Erwin. Results of the comparison are in Table \ref{tab:all_evaluations}.}
    \label{tab:hyperparameters_comparison}
\end{table}

\section{Hyperparameters for reproducibility experiments}
\label{sec:appendix reproduce}
\begin{table}[H]
    \centering
    \begin{tabular}{@{}llc@{}}
\toprule
Dataset       & Hyperparameter            & Value            \\ \midrule
\multirow{4}{*}{Cosmology} 
              & Batch Size                 & 16 \\
              & Ball Size                  & 256 \\
              & Number of Epochs           & 5000 \\
              & Number of Training Samples & 8192 \\
              \midrule
\multirow{3}{*}{MD} 
              & Batch Size                 & 32 \\
              & Number of Epochs           & 50000 \\
              & Number of trainable parameters & 4M \\
              \midrule
\multirow{3}{*}{ShapeNet} 
              & Batch Size                 & 2 \\
              & Ball Size                  & 256 \\
              & Number of Epochs           & 1000 \\
              \bottomrule
\end{tabular}
    \caption{Summary of hyperparameters used for reproducibility experiments. We used the default "small" configuration of Erwin. The parameters in the table are critical for replication, while the absent ones are set to the same values as in the original paper's appendix. Results of the comparison are in Table \ref{table:replicate}.}
\end{table}
\end{document}